\documentclass[lettersize,journal]{IEEEtran}
\usepackage{amsmath,amsfonts}
\usepackage{algorithmic}
\usepackage{array}
\usepackage[caption=false,font=normalsize,labelfont=sf,textfont=sf]{subfig}
\usepackage{textcomp}
\usepackage{stfloats}
\usepackage{url}
\usepackage{verbatim}
\usepackage{graphicx}
\usepackage{booktabs,graphicx,xcolor,multirow}
\usepackage{arydshln}
\def\BibTeX{{\rm B\kern-.05em{\sc i\kern-.025em b}\kern-.08em
    T\kern-.1667em\lower.7ex\hbox{E}\kern-.125emX}}
\usepackage{balance}
\begin{document}
\title{AMIF: Authorizable Medical Image Fusion Model with Built-in Authentication}
\author{Jie Song\textsuperscript{1}, Jun Jia\textsuperscript{2}*, Wei Sun\textsuperscript{3}, Wangqiu Zhou\textsuperscript{4}, Yue Sun\textsuperscript{1}, Tao Tan\textsuperscript{1}*,  Guangtao Zhai\textsuperscript{2}*
\thanks{This work was supported by the Shenzhen Medical Research Fund under Grant No. A2603012, the Science and Technology Development Fund of Macao under Grant No. 0014/2025/RIC, Macao Polytechnic University under Grant No. RP/FCA-17/2025, and the National Natural Science Foundation of China under Grant No. 62601018. \\
\indent *Corresponding authors:Jun Jia, Tao Tan, Guangtao Zhai.\\
\indent  \textsuperscript{1}Jie Song, Yue Sun, Tao Tan are with the Faculty of Applied Sciences, Macao Polytechnic University, Macau 999078, China (e-mail: p2316172@mpu.edu.mo; yuesun@mpu.edu.mo; taotanjs@gmail.com). \\
\indent \textsuperscript{2}Jun Jia and Guangtao Zhai are with the School of Electronic Information and Electrical Engineering, Shanghai Jiao Tong University, Shanghai, 200240, China (e-mail: jiajun0302@sjtu.edu.cn; zhaiguangtao@sjtu.edu.cn).\\
\indent \textsuperscript{3}Wei Sun is with the School of Information and Electronic Engineering, East China Normal University, Shanghai, 200241, China (e-mail: wsun@cee.ecnu.edu.cn).\\
\indent \textsuperscript{4}Wangqiu Zhou is with the School of Computer Science and Information Engineering, Hefei University of Technology, Hefei, 230009, China (e-mail: rafazwq@hfut.edu.cn).
}}

\maketitle

\begin{abstract}
Multimodal medical image fusion integrates complementary anatomical and functional information from different imaging modalities, providing richer image representations for clinical interpretation and assisted diagnosis. High-performance medical image fusion models are typically built upon medical images with high data sensitivity and security requirements, together with carefully designed algorithmic architectures. Collectively, these constitute intellectual property of significant value. As fusion results are circulated and reused in scenarios such as cross-institutional collaboration, traceable copyright protection and controlled authorization for these medical fusion assets are becoming increasingly important. However, existing multimodal medical image fusion models typically generate unprotected results and lack built-in copyright protection and authorization mechanisms. Conventional post-processing watermarking is independent of the fusion model, making it difficult for copyright protection to become an intrinsic capability of the model itself. To address this issue, we propose \textbf{AMIF}, the first \textbf{A}uthorizable \textbf{M}edical \textbf{I}mage \textbf{F}usion model with built-in authentication, which integrates authorization access control into the image fusion objective. For unauthorized usage, AMIF embeds explicit and visible copyright identifiers into fusion results. In contrast, high-quality fusion results are accessible upon successful key-based authentication. To maintain fusion quality for authorized parties and enhance robustness to watermark removal, AMIF exploits Content-Conditioned Watermark Memory (CCWM) and Channel–Spatial Attention Modulated Invertible Coupling (C-SAMIC) mechanisms that shift watermark embedding from an external input-dependent mechanism to a model-internal capability. Extensive experiments demonstrate that AMIF achieves flexible copyright protection while exhibiting strong robustness against watermark removal attacks.
\end{abstract}
\vspace{-4pt}
\begin{IEEEkeywords}
Multimodal Medical Image Fusion, Authorizable, Copyright Protection
\end{IEEEkeywords}
\vspace{-16pt}
\section{Introduction}
\IEEEPARstart{M}{ultimodal} image fusion (MMIF) has been widely studied and applied in medical imaging \cite{zhou2023deep,zhao2023cddfuse}. In medical imaging, different modalities emphasize different aspects of anatomy and physiology. For example, Computed Tomography (CT) excels at depicting bones and other high-density structures. Magnetic Resonance Imaging (MRI) provides superior soft-tissue contrast and lesion delineation. Functional modalities such as Positron Emission Tomography/Single-Photon Emission Computed Tomography (PET/SPECT) offer cues about metabolic activity and perfusion. Fusing these complementary cues yields a more information-rich visual representation. It can improve downstream tasks such as tumor segmentation\cite{zhang2024robust}. It can also better support clinical diagnosis and treatment planning\cite{james2014medical,zhang2024important}. To improve fusion quality and downstream task performance, great efforts have been made in the network architecture design\cite{IFCNN,SDNet,U2Fusion}, feature interaction\cite{SwinFusion}, fusion strategies\cite{PSLPT}, and loss-function optimization\cite{ref52}. As medical image fusion gradually moves toward practical applications, the development of professional medical image fusion models becomes a collaborative effort supported by multiple forms of specialized resources and expertise. It relies not only on hospital medical images acquired under standardized procedures and characterized by high data sensitivity and security requirements, but also on domain expertise from radiologists and clinicians, as well as algorithmic design capabilities from medical imaging AI researchers. Therefore, these professional medical image fusion models, jointly supported by medical imaging resources, clinical expertise, and algorithmic design, carry significant intellectual property value and require appropriate protection.
\\
\begin{figure}
    \centering
    \includegraphics[width=0.95\linewidth]{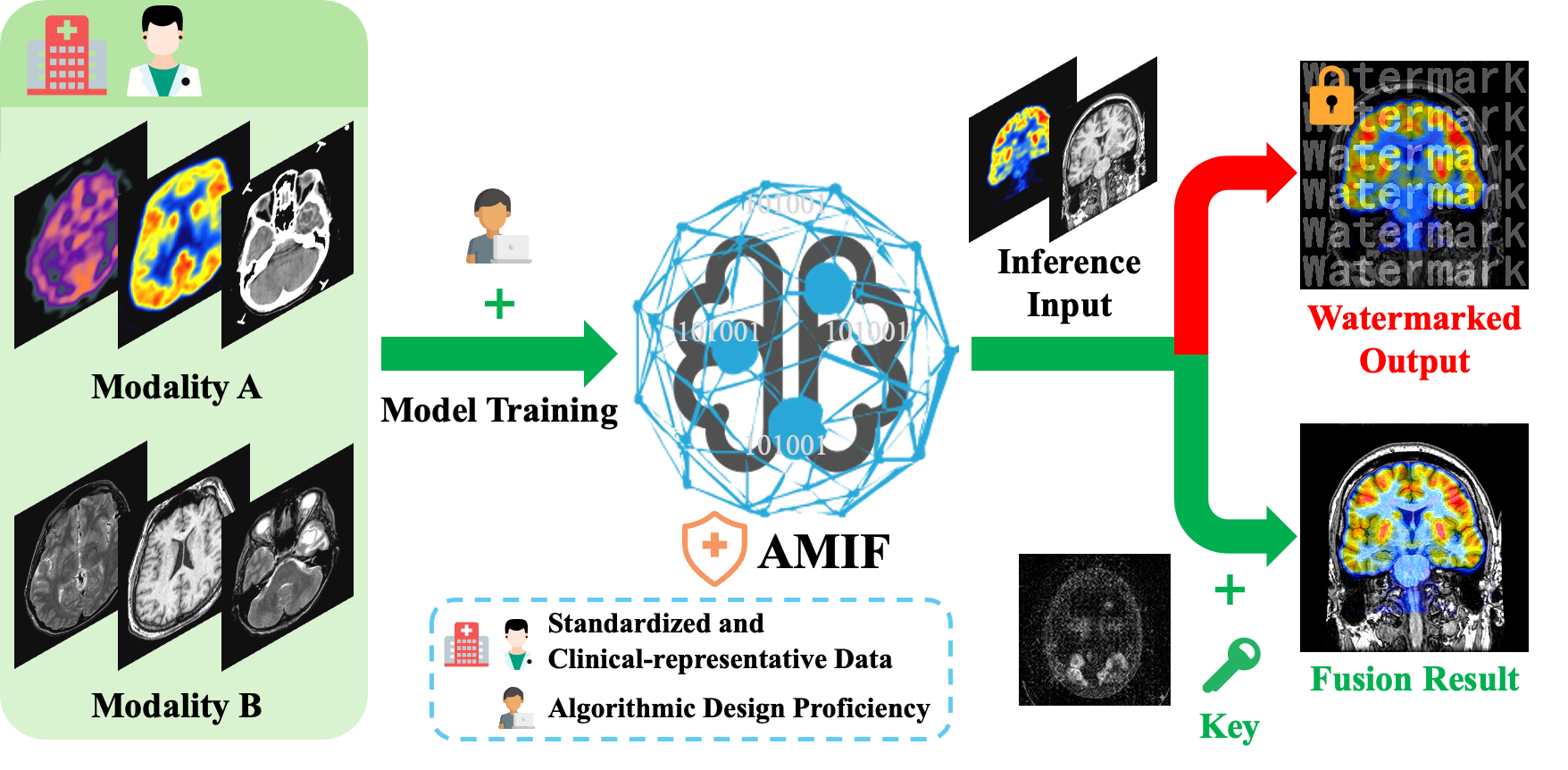}
    \caption{Why AMIF? AMIF embeds copyright protection and authorization into the medical image fusion process. It produces a visibly watermarked fusion result by default, while a valid key enables authorized recovery of a high-quality watermark-free result. This built-in design couples fusion, copyright protection, and authorized recovery within a unified model.}
    \label{fig:application}
    \vspace{-15pt}
\end{figure}
\indent However, current medical image fusion models largely overlook intellectual property (IP) protection during deployment and clinical dissemination. With the growing circulation of fused medical images through telemedicine, cross-institutional collaboration, and open research repositories, these fusion results are increasingly exposed to unauthorized use, redistribution, tampering, and IP infringement. Existing fusion models usually output “bare” fused results. They lack built-in ownership verification and copyright protection. Post-hoc watermarking is a possible remedy. Yet it often relies on external watermark models or third-party services. This can introduce data leakage risks. It also separates “fusion” from “protection” into two independent stages. As a result, the watermark is weakly coupled with the fused content. It becomes easier to remove under attacks. In practice, real-world deployment also requires authorization. Authorized users should be able to remove the watermark losslessly before downstream analysis or clinical diagnosis. This should be enabled by an authorization key. Therefore, a unified model is required to achieve high-quality fusion, enable traceable copyright protection, and support revocable authorization via controllable watermark removal. This is crucial for secure and compliant use of medical images in real-world circulation.
Therefore, we propose \textbf{AMIF}, the first \textbf{A}uthorizable \textbf{M}edical \textbf{I}mage \textbf{F}usion model with built-in authentication. AMIF jointly formulates copyright protection constraints and the fusion objective in a unified framework for the first time. AMIF further introduces a key-based authorization mechanism to meet the strict requirements of clinical integrity and usability. Fig.~\ref{fig:application} shows the end-to-end application workflow of AMIF. In the unauthorized setting, the model outputs fused images with visible copyright watermarks, which are suitable for public sharing and storage. Given the correct key, the model can controllably remove the watermark and recover a clean, watermark-free fused result. This supports authorized clinical usage and downstream analysis. To maintain the fusion quality for authorized usage, we design a Content-Conditioned Watermark Memory (CCWM) module. It enables the model to generate watermark representations from internal parameters, without requiring explicit watermark inputs. CCWM introduces a set of learnable watermark memory vectors. It uses bidirectional cross-attention to couple the watermark memory with input features. This builds a content-binding relationship between the watermark representation and the input features. It facilitates tighter writing of watermark signals into the protected representation. Furthermore, to enhance robustness against watermark removal attacks, we propose a Channel–Spatial Attention Modulated Invertible Coupling (C-SAMIC) mechanism. It enhances the integration between image representations and watermark signals at the feature level. This improves the compatibility between the watermark and image features. It also strengthens robustness against unauthorized removal operations. In summary, our contributions are as follows:
\vspace{-2pt}
\begin{itemize}
	\item We propose \textbf{AMIF}, the first \textbf{A}uthorizable \textbf{M}edical \textbf{I}mage \textbf{F}usion model with built-in authentication. AMIF explicitly introduces copyright protection into the fusion objective while enabling high-quality fusion results accessible for authorized usage;
	\item We design the Content-Conditioned Watermark Memory (CCWM) module and Channel–Spatial Attention Modulated Invertible Coupling (C-SAMIC) mechanism to enhance authorized fusion quality and watermark-removal robustness, respectively;
	\item Experiments on public medical datasets show that AMIF can controllably remove the watermark and recover a watermark-free result under valid key authorization. It also achieves competitive fusion performance compared to existing State of the Art (SOTA) methods.
\end{itemize}
\vspace{-12pt}
\section{Related Works}
\subsection{Deep Learning-based Multimodal Image Fusion}
Deep learning-based multimodal image fusion has become a mainstream research direction. Early methods are mostly built on Convolutional Neural Networks (CNNs), which learn modality-specific representations and fuse them in the feature space to improve fusion quality\cite{IFCNN,SDNet,U2Fusion,ref52,deng2020deep,gao2022multi}. Beyond CNN frameworks, some methods use Generative Adversarial Networks (GANs) to learn distribution mappings for fusion\cite{li2020attentionfgan,zhang2023transformer}. In recent years, Transformer-based methods employ attention to exchange cross-modal information and model long-range dependencies, improving the representation of structures and texture details\cite{SwinFusion}. In addition, some works explore Mamba-based selective state-space modeling for fusion\cite{xie2024fusionmamba,peng2024fusionmamba}. For example, Xie et al.\cite{xie2024fusionmamba} propose a dynamic feature fusion module to enhance dynamic textures, difference awareness, and cross-modal feature strengthening, while suppressing redundant information. On the other hand, Wang et al.\cite{PSLPT} design fusion strategies from multi-scale or frequency-domain perspectives to better preserve salient structures and details. Zhao et al.\cite{zhao2023cddfuse} propose a correlation-driven feature decomposition fusion network, with a correlation-driven loss that encourages low-frequency correlation and high-frequency complementarity. Liang et al.\cite{liang2022fusion} propose DeFusion, which performs fusion via decomposition-based representation learning. To address cross-modal misalignment, Huang et al.\cite{huang2022reconet} propose ReCoNet, which uses lightweight registration and recurrent correction to alleviate artifacts and improve robustness. Nam et al.\cite{nam2022neural} use implicit coordinate-based neural representations for multi-image alignment and fusion. In addition, image fusion is often combined with downstream tasks such as segmentation, where more informative fused representations can improve segmentation accuracy\cite{liu2023multi,maiti2023transfusion,zhang2024robust}.
\vspace{-10pt}
\subsection{Copyright Protection and Image Watermarking}
Image watermarking is a widely used technique for copyright protection. It embeds ownership information into images to support attribution and traceability. By presentation, watermarks can be visible\cite{niu2023fine,zhang2025dvw} or invisible\cite{zhang2021deep}. Traditional watermarking methods typically rely on hand-crafted signal processing and embedding rules in transform domains\cite{barni2001improved,hernandez2000dct}. Recently, deep learning-based watermarking methods\cite{tancik2020stegastamp,bui2023rosteals} have further improved the flexibility of watermark embedding. However, most existing watermarking pipelines are used only as a post-processing step. They are decoupled from upstream generative tasks such as image fusion. This often leads to weak coupling between the watermark signal and image content. In this work, we unify watermark-based copyright protection and multimodal medical image fusion in a single modeling framework.

\begin{figure*}
    \centering
    \includegraphics[width=0.90\linewidth]{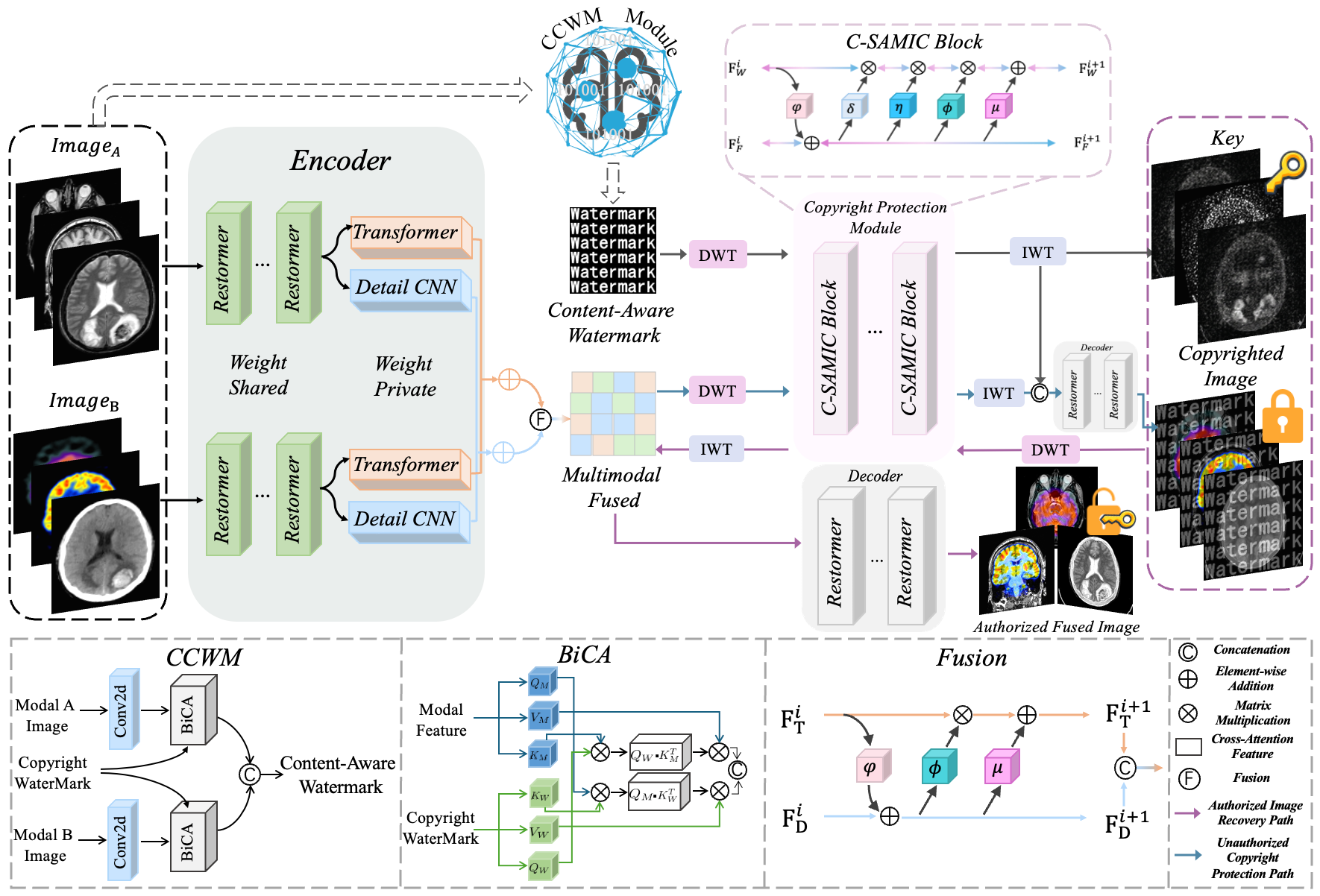}
    \caption{AMIF enables authorizable multimodal medical image fusion with built-in authentication. A shared Restormer encoder extracts common features and private encoders capture modality-specific cues, which are fused into a multimodal representation. CCWM first generates an internal content-aware watermark that carries implicit content cues. C-SAMIC then injects it into the protected features via tightly-coupled wavelet-domain invertible coupling, producing a copyrighted output and an associated key. With a valid key, the inverse mapping removes the watermark and recovers a high-quality watermark-free fusion result for authorized use. The whole pipeline works in a unified and cooperative manner.}
    \label{fig:OverAll}
\end{figure*}
\vspace{-12pt}
\subsection{Invertible Neural Networks}
The concept of invertible neural networks traces back to Non-linear Independent Components Estimation (NICE) introduced by Dinh et al.\cite{dinh2014nice}, which adopts invertible additive coupling transforms to realize bidirectional mappings. Building on this idea, Dinh et al.\cite{dinh2016density} further introduce convolutional structures into coupling models and use a multi-scale design to reduce computational cost. Kingma et al.\cite{kingma2018glow} incorporate key components such as invertible convolutions, improving model expressiveness and scalability and making INN more suitable for image modeling. Subsequent studies extend INN to a range of vision tasks, including deep image hiding\cite{xu2022robust}, adversarial example generation\cite{chen2023imperceptible}, super-resolution\cite{xiao2023invertible}, and improving backbone feature representations for image classification\cite{behrmann2019invertible,gomez2017reversible}.
To embed watermark signals effectively in image features and recover high-quality watermark-free fusion results under authorization, we tailor an invertible framework to strengthen the binding between watermark identifiers and image representations.
\vspace{-13pt}
\section{Method}
\subsection{Overall Architecture}
In practical medical image fusion scenarios, fusion results require explicit copyright protection during sharing, circulation, and use, while authorized users should still be able to obtain high-quality watermark-free fusion results. Therefore, copyright protection should not be treated merely as an external post-processing step after fusion, but should be built into the medical image fusion model itself, making copyright protection an intrinsic capability of the model and enabling intellectual property protection for both the fusion model and its generated results. Motivated by this requirement, we propose an authorizable medical image fusion framework, AMIF, as shown in Fig.~\ref{fig:OverAll}. AMIF integrates multimodal information fusion, copyright-protected output generation, and key-based authorized recovery within a unified framework.

Given two input images from different modalities, $Image_A$ and $Image_B$, we use Restormer\cite{zamir2022restormer} as the shared encoder and decoder to extract cross-modal base representations. We then introduce a private encoder for each modality to capture modality-specific features. Each private encoder contains a global branch based on Lite Transformer\cite{wu2020lite} and a local Detail CNN branch based on INN\cite{dinh2016density}. This design models both long-range dependencies and fine textures. Features from the two modalities are integrated by a fusion module that consists of the $\varphi$, $\phi$, and $\mu$ functions to obtain a fused multimodal feature. In the unauthorized mode, the CCWM module generates a watermark representation from internal learnable memory parameters and conditions it on the input images, as detailed in Sec.~3.2. Because the watermark interacts with the input features during generation, it implicitly encodes content cues from the current multimodal pair. This makes the watermark representation more compatible with the fused content. It also facilitates subsequent writing into protected features and stabilizes the later key-based recovery. We then map the fused content representation and the CCWM watermark representation to the wavelet domain. We apply the discrete wavelet transform (DWT) to decompose them into multi-frequency components. This frequency decomposition allows us to inject watermark signals in a more controllable manner across different bands. Next, we concatenate the wavelet-domain features and feed them into the Copyright Protection Module. This module consists of $N$ C-SAMIC blocks (as detailed in Sec.~3.3). Each C-SAMIC block performs an invertible feature coupling and progressively writes watermark cues into the content stream at the feature level.

Importantly, the copyright protection module produces two outputs: the protected fused features and a feature stream stripped of redundant information that serves as the key for inversion. The protected features are decoded by the Restormer decoder to produce the copyrighted fused image. In the authorized mode, given a valid key, the model applies the inverse mapping of the C-SAMIC blocks to remove the watermark in feature space and reconstruct an authorized fused image.
\vspace{-9pt}
\subsection{CCWM Module}
Existing medical image fusion models generally provide limited support for copyright protection. When conventional watermarking is applied, a fixed watermark is typically embedded as an additional step after fusion. Such an external watermark can be tampered with, replaced, or bypassed, thereby weakening the protection, while its representation remains independent of the source-image content. Since the fusion result is jointly formed from the current multimodal inputs, a fixed watermark is difficult to adapt to the specific fusion content and may limit its compatibility with the fused features. To address these limitations, we propose CCWM, which transforms the copyright watermark from an external input into a learnable internal representation and further conditions it on the source-image content. Consequently, the watermark can extract and encode content-relevant information from the input images while retaining implicit cross-modal cues to support subsequent image recovery and improve its compatibility with the fused features during embedding. CCWM initializes a set of learnable watermark memory vectors as the carrier of copyright identifiers. Lightweight 2D convolutions extract features from the two input modalities. Next, Bidirectional Cross Attention (BiCA) builds a two-way association between the watermark vectors and the input features. On one hand, the watermark vectors attend to the input features to form content-conditioned watermark features. On the other hand, the input features attend back to the watermark vectors, strengthening their mutual awareness. Finally, we concatenate the two watermark features along the channel dimension to obtain the final content-aware watermark representation.
\vspace{-10pt}
\subsection{C-SAMIC Module}
In copyright-protected medical image fusion, watermark embedding needs to satisfy two competing requirements: on the one hand, the copyright information should be sufficiently coupled with the fused features to resist unauthorized watermark removal; on the other hand, the coupling should remain reversible to ensure high-quality watermark-free recovery for authorized users. A simple one-way embedding strategy is difficult to satisfy both requirements. To this end, we design the C-SAMIC module, which establishes bidirectional interactions between the fused features and watermark features through invertible coupling, while channel and spatial attention modulate the coupling along the channel and spatial dimensions, respectively. Specifically, we split the input into two parts: $F_F^i$ denotes the content features to be protected, and $F_W^i$ denotes the watermark features. Each coupling block consists of a set of learnable functions, denoted as $\varphi$, $\delta$, $\eta$, $\phi$, and $\mu$. The functions $\varphi$, $\phi$, and $\mu$ are implemented with dense blocks~\cite{wang2018esrgan}. $\delta$ adopts the channel attention mechanism in~\cite{hu2018squeeze} to strengthen information flow exchange along the channel dimension. $\eta$ introduces a spatial attention mechanism to guide feature interaction and alignment across spatial locations. In the unauthorized mode, the copyright protection is computed as follows:
\vspace{-7pt}
\begingroup
\small
\begin{align}
F_{\mathrm{F}}^{i+1}
&= F_{\mathrm{F}}^{i}
+ \varphi\!\left(F_W^{i}\right),\\
F_W^{i+1}
&= F_W^{i}
\odot \delta\!\left(F_{\mathrm{F}}^{i+1}\right)
\notag\\
&\quad \odot
\exp\!\Bigl(
\alpha\bigl(
\eta(F_{\mathrm{F}}^{i+1})
\odot \phi(F_{\mathrm{F}}^{i+1})
\bigr)
\Bigr)
+ \mu(F_{\mathrm{F}}^{i+1}),
\end{align}
\endgroup
where $\alpha(\cdot)$ denotes a sigmoid function scaled by a constant factor, and $\odot$ denotes element-wise multiplication.\\
In the authorized mode, watermark recovery corresponds to the inverse computation of the copyright protection process, with the information flow in the opposite direction. The formulation is given as follows:
\begingroup
\small
\begin{align}
F_K^{i}
&= F_K^{i+1}
- \mu\!\left(F_{\mathrm{C}}^{i+1}\right)
\notag\\
&\quad \odot
\exp\!\Bigl(
-\alpha\bigl(
\phi(F_{\mathrm{C}}^{i+1})
\oslash \eta(F_{\mathrm{C}}^{i+1})
\bigr)
\Bigr)
\oslash \delta(F_{\mathrm{C}}^{i+1}),\\
F_{\mathrm{C}}^{i}
&= F_{\mathrm{C}}^{i+1}
- \varphi\!\left(F_K^{i}\right).
\end{align}
\endgroup
where $F_K^i$ denotes the key, $F_C^i$ denotes the copyrighted image, and $\oslash$ denotes element-wise division.
\vspace{-15pt}
\subsection{Loss Function}
AMIF is trained to align with two usage modes. In the unauthorized mode, the forward pass produces a watermarked fused output. In the authorized mode, the inverse pass removes the watermark under a given key and recovers a watermark-free result. Accordingly, we group the training objective into two parts. The first part enforces fusion quality and watermark embedding in the unauthorized mode. The second part enforces invertible recovery quality in the authorized mode. We describe these two loss terms in detail below.\\
\textbf{Authorized mode loss.}
In the authorized mode, AMIF needs to generate watermark-free results to support clinical applications and downstream analysis. Inspired by works such as \cite{tang2022image,zhao2023cddfuse}, we design the loss function by considering intensity fidelity, gradient consistency, feature decomposition constraints, and key-conditioned feature recovery, as defined below.
\vspace{-6pt}
\begin{equation}
\mathcal{L}_{\text{fusion}}
=
\mathcal{L}_{\text{int}}
+
\alpha_1 \mathcal{L}_{\text{grad}}
+
\alpha_2 \mathcal{L}_{\text{decomp}}
+
\alpha_3 \mathcal{L}_{\text{krecov}}
\label{eq:authorizedloss}
\end{equation}
\vspace{-13pt}
\begin{equation}
\mathcal{L}_{\text{krecov}}
=
\frac{1}{N}\sum_{n=1}^{N}\left(F_{\text{recov}}^{(n)}-F_{\text{ori}}^{(n)}\right)^2
\label{eq:keyrecovloss}
\end{equation}
\indent The detailed formulations of the intensity, gradient, and
decomposition losses are provided in Supplementary
Materials (S.I-A). $\alpha_1$, $\alpha_2$, and $\alpha_3$ are tuning parameters.\\
\textbf{Unauthorized mode loss.}
In the unauthorized mode, AMIF generates a watermark representation from learnable built-in parameters and writes it into the fused result. The output thus carries a verifiable copyright identifier. To supervise the learning of these internal watermark parameters, we align the predicted watermark with a predefined watermark label and optimize it with a combined Binary Cross-Entropy (BCE) and Dice loss. Specifically, BCE provides stable pixel-wise supervision, while Dice mitigates class imbalance caused by the sparse watermark region and better constrains the watermark shape and structure. In addition, to write the watermark signal into the protected features in a more natural and stable way, we introduce an MSE-based content consistency constraint. Specifically, we enforce an MSE loss in the pixel domain between the watermarked fused output $I_{\mathrm{wf}}$ and a reference image $I_{\mathrm{wft}}$. The reference image is defined as the average of the two input modalities and the watermark label, i.e., $\tfrac{1}{3}(I_{\mathrm{modalA}}+I_{\mathrm{modalB}}+I_{\mathrm{wt}})$. Moreover, to further regularize structural consistency, we map features to the wavelet domain and compute an MSE loss on the low-frequency components of DWT. This encourages stable watermark injection in the low-frequency structural band. The detailed formulations are provided in S.I-B of the supplementary materials. Finally, the overall optimization loss function for AMIF can be summarized as:
\begin{equation}
\mathcal{L}_{\text{total}}
=
\mathcal{L}_{\text{fusion}}
+
\alpha_4((\mathcal{L}_{\text{wm}}
+
\mathcal{L}_{\text{wmlow}})
+
\alpha_5(\mathcal{L}_{\text{BCE}}
+
\mathcal{L}_{\text{Dice}}))
\label{eq:total}
\end{equation}
\vspace{-27pt}
\section{Experiments}
\subsection{Experiment Setting}
\textbf{Datasets and metrics.} To evaluate the proposed AMIF model, we conduct experiments on public medical images from the Harvard medical image website\footnote{\url{http://www.med.harvard.edu/AANLIB/home.html}}. The dataset covers three typical multimodal pairs, including MRI-CT, MRI-PET, and MRI-SPECT. Since the MRI-CT subset is relatively small, we apply random rotation for data augmentation to increase its size. We merge all pairs into a unified dataset, with 983 pairs for training, 104 pairs for validation, and 81 pairs for testing. The test set includes 16 MRI-CT pairs, 25 MRI-PET pairs, and 40 MRI-SPECT pairs. For quantitative evaluation, we follow~\cite{ma2019infrared} and report four common fusion metrics. Mutual information (MI) quantifies how much information from the source images is preserved in the fused result. Visual information fidelity (VIF) evaluates perceptual fidelity by measuring how well visual information is retained. $Q^{AB/F}$ assesses edge information transfer from the sources to the fused image. Structural similarity (SSIM) measures structural consistency between images. Higher values generally indicate better fusion quality.\\
\indent \textbf{Implement details.} All experiments are conducted on a server equipped with an NVIDIA GeForce RTX A6000 GPU (48GB memory). 
All images are resized to $256\times256$. 
We train for 200 epochs with a batch size of 2. We use the Adam optimizer with an initial learning rate of $1\times10^{-4}$, which is decayed by a factor of 0.5 every 100 epochs. 
In the encoder, both Restormer and the Transformer use 4 blocks, with 8 attention heads and 64 feature dimensions. 
The decoder follows the same configuration. 
For Eqs.~\eqref{eq:authorizedloss} and \eqref{eq:total}, we set $\alpha_1$ to $\alpha_5$ to 10, 2, 100, 0.1, and 0.1 to match the intended optimization priorities.
\vspace{-5pt}
\subsection{Authorization-aware Fusion and Copyright Protection Analysis}

\begin{figure}
	\centering
	\includegraphics[width=1\linewidth]{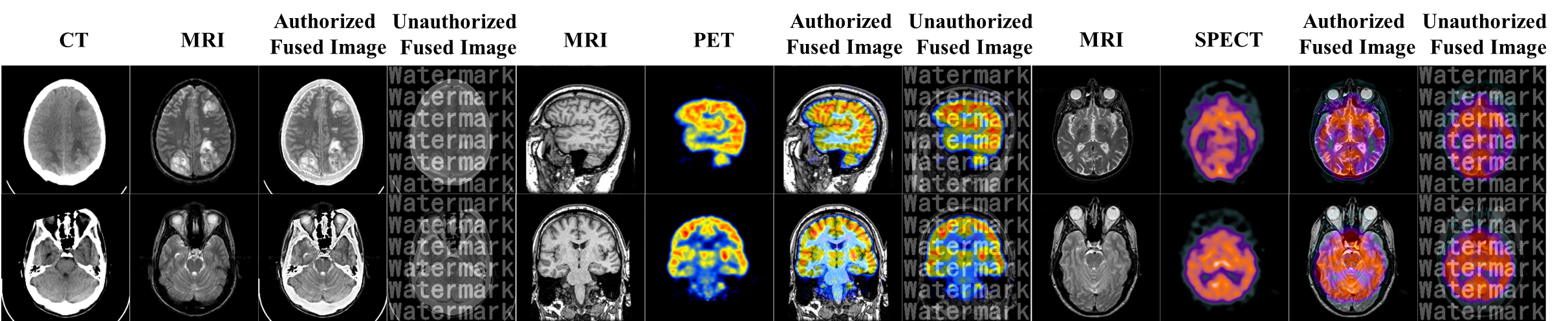}
	\caption{Qualitative fusion results of three modality combinations (CT-MRI, MRI-PET, and MRI-SPECT) under unauthorized and authorized modes. The unauthorized mode generates fused images with copyright watermarks, while the authorized mode generates fused images recovered using the key.}
	\label{fig:auunaushow}
    \vspace{-15pt}
\end{figure}

In this subsection, we conduct qualitative and quantitative experiments to examine the behavior of AMIF under unauthorized and authorized use. \\
\textbf{Quantitative Analysis.}
As reported in Table~\ref{table:Aufusionquantresults}, AMIF exhibits clear quantitative differences between the authorized and unauthorized settings. Under unauthorized use, VIF, $Q^{AB/F}$, and SSIM decrease substantially across all three modality combinations, indicating that the visible copyright watermark has a pronounced influence on perceptual quality, edge preservation, and structural fidelity. In contrast, MI shows a more moderate decrease because the protected outputs still retain part of the information shared with the source modalities despite the copyright watermark. After authorization, all four metrics improve substantially, and the recovered results achieve competitive fusion performance. These results further confirm that AMIF maintains a clear distinction between copyright-protected and authorized outputs while preserving high-quality fusion performance after valid key-based recovery. \\
\textbf{Qualitative Analysis.}
As shown in Fig.~\ref{fig:auunaushow}, AMIF produces visibly watermarked fusion results by default in the unauthorized setting. The copyright identifiers remain clear and readable across different modality combinations, providing an explicit ownership declaration. Meanwhile, the protected outputs preserve the main anatomical structures and complementary information from the source modalities without obvious large-scale structural distortions. After a valid key is provided, AMIF removes the visible copyright identifiers and recovers watermark-free fusion results. Compared with the unauthorized outputs, the authorized results exhibit cleaner local regions, clearer texture details, and sharper structural boundaries. These observations demonstrate the intended authorization mechanism of AMIF: unauthorized outputs retain the main fusion information while remaining explicitly copyright-protected, whereas high-quality watermark-free results can only be recovered using a valid key.
\begin{table}[t]
    \vspace{-15pt}
    \centering
    \caption{Quantitative comparison of AMIF under authorized and unauthorized scenarios. The authorized results are compared with state-of-the-art fusion methods after watermark removal using a valid key. The unauthorized results are reported to demonstrate the impact of copyright watermark embedding. (\textcolor{red}{Red} indicates the best results, and \textcolor{blue}{Blue} indicates the second best).}
    \label{table:Aufusionquantresults}

    \begingroup
    \scriptsize
    \setlength{\tabcolsep}{1.2pt}
    \renewcommand{\arraystretch}{0.90}

    \resizebox{\columnwidth}{!}{%
        \begin{tabular}{@{}lcccc cccc cccc@{}}
            \toprule
            \bfseries Methods
            & \multicolumn{4}{c}{\bfseries MRI-CT}
            & \multicolumn{4}{c}{\bfseries MRI-PET}
            & \multicolumn{4}{c}{\bfseries MRI-SPECT} \\
            \cmidrule(lr){2-5}\cmidrule(lr){6-9}\cmidrule(lr){10-13}

            & \bfseries MI $\uparrow$
            & \bfseries VIF $\uparrow$
            & $\mathbf{Q^{AB/F}}$ $\uparrow$
            & \bfseries SSIM $\uparrow$

            & \bfseries MI $\uparrow$
            & \bfseries VIF $\uparrow$
            & $\mathbf{Q^{AB/F}}$ $\uparrow$
            & \bfseries SSIM $\uparrow$

            & \bfseries MI $\uparrow$
            & \bfseries VIF $\uparrow$
            & $\mathbf{Q^{AB/F}}$ $\uparrow$
            & \bfseries SSIM $\uparrow$ \\
            \midrule

            IFCNN\cite{IFCNN}
            & 1.81 & 0.40 & 0.55 & 0.65
            & 1.63 & 0.44 & 0.55 & 0.43
            & 1.56 & 0.46 & 0.57 & 0.56 \\

            U2FUSION\cite{U2Fusion}
            & 1.66 & 0.37 & 0.40 & 0.44
            & 1.76 & 0.43 & 0.38 & 0.34
            & 1.61 & 0.43 & 0.37 & 0.39 \\

            SWINFUSION\cite{SwinFusion}
            & 2.01 & 0.56 & \textcolor{blue}{0.59} & \textcolor{blue}{0.69}
            & \textcolor{blue}{2.63} & \textcolor{blue}{0.79} & \textcolor{blue}{0.74} & \textcolor{red}{0.57}
            & \textcolor{blue}{2.23} & \textcolor{blue}{0.84} & \textcolor{blue}{0.76} & \textcolor{red}{0.64} \\

            PSLPT\cite{PSLPT}
            & 1.62 & 0.37 & 0.35 & 0.50
            & 1.74 & 0.51 & 0.39 & 0.38
            & 1.63 & 0.53 & 0.41 & 0.46 \\

            FUSIONMAMBA\cite{xie2024fusionmamba}
            & 2.01 & \textcolor{blue}{0.57} & 0.52 & \textcolor{red}{0.70}
            & 2.01 & 0.59 & 0.62 & 0.47
            & 1.93 & 0.69 & 0.70 & \textcolor{blue}{0.62} \\

            MMIF-INET\cite{he2025mmif}
            & 1.74 & 0.48 & 0.58 & 0.62
            & 1.90 & 0.57 & 0.64 & 0.44
            & 1.92 & 0.74 & 0.74 & 0.58 \\

            DM-FNET~\cite{DM-FNET}
            & \textcolor{blue}{2.03} & 0.56 & 0.58 & \textcolor{blue}{0.69}
            & 2.07 & 0.62 & 0.69 & \textcolor{blue}{0.54}
            & 1.84 & 0.63 & 0.69 & 0.52 \\

            OURS (Authorized)
            & \textcolor{red}{2.04} & \textcolor{red}{0.59} & \textcolor{red}{0.64} & \textcolor{red}{0.70}
            & \textcolor{red}{2.97} & \textcolor{red}{0.85} & \textcolor{red}{0.76} & \textcolor{red}{0.57}
            & \textcolor{red}{2.68} & \textcolor{red}{0.93} & \textcolor{red}{0.78} & \textcolor{red}{0.64} \\

            \hdashline

            OURS (Unauthorized)
            & 1.62 & 0.09 & 0.12 & 0.19
            & 1.80 & 0.12 & 0.13 & 0.15
            & 1.69 & 0.09 & 0.11 & 0.16 \\

            \bottomrule
        \end{tabular}%
    }
    \endgroup
    \vspace{-15pt}
\end{table}
\begin{figure*}
    \centering
\includegraphics[width=0.9\linewidth]{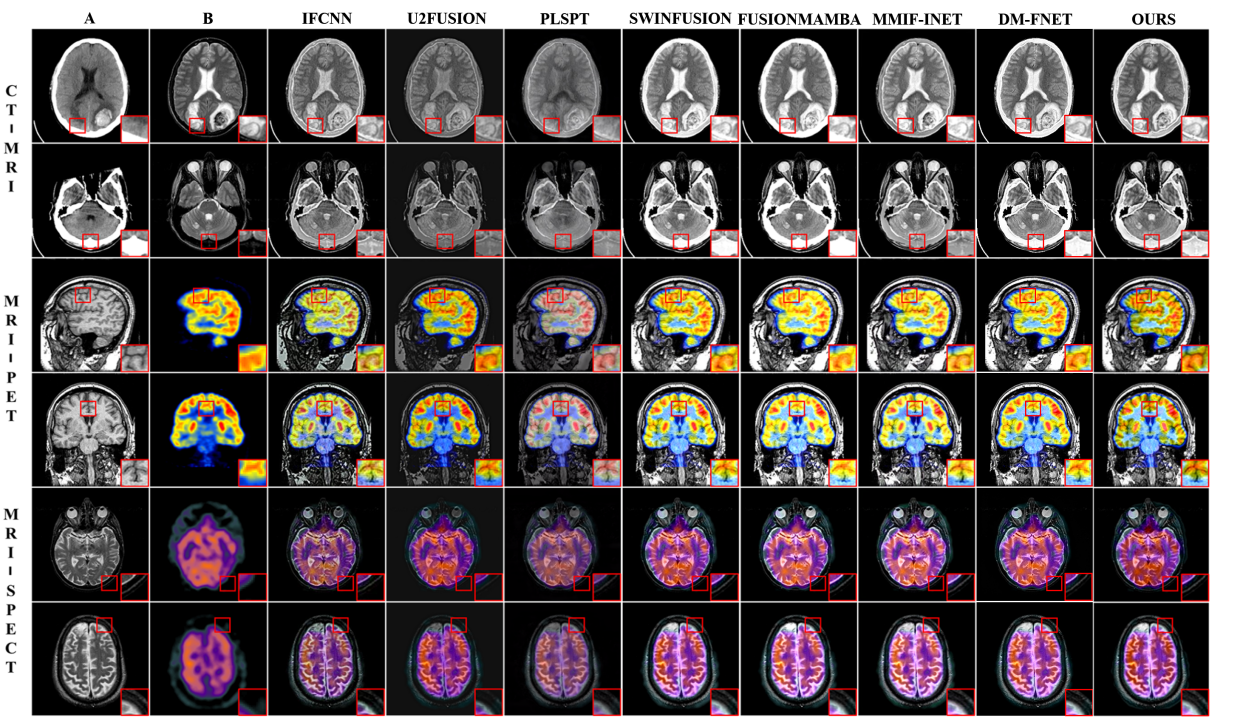}
    \caption{Qualitative results of authorized (watermark-free) fusion images produced by AMIF compared with different fusion models on three multimodal datasets. The first column A and the second column B respectively represent two types of input modalities.}
    \label{fig:AuthorizedFusionResults}
\end{figure*}
\vspace{-12pt}
\subsection{State-of-the-Art Comparison on Authorized Fusion}
In this section, we evaluate AMIF on the test set in the authorized mode, where the watermark is effectively removed. We compare AMIF with state-of-the-art methods using both quantitative metrics and qualitative results, including IFCNN~\cite{IFCNN}, U2FUSION~\cite{U2Fusion}, FUSIONMAMBA~\cite{xie2024fusionmamba}, PSLPT~\cite{PSLPT}, SWINFUSION~\cite{SwinFusion}, DM-FNET~\cite{DM-FNET} and MMIF-INET~\cite{he2025mmif}.
\\
\textbf{Quantitative Analysis.}
As shown in Table~\ref{table:Aufusionquantresults}, AMIF achieves the best MI, VIF, and $Q^{AB/F}$ on all three modality combinations, while its SSIM is tied with the best-performing competing methods. For MRI--CT fusion, IFCNN, U2FUSION, and PSLPT obtain relatively lower VIF and $Q^{AB/F}$ values, indicating weaker preservation of perceptual information and edge details. MMIF-INET, SWINFUSION, and DM-FNET achieve improved edge preservation, with their $Q^{AB/F}$ values ranging from 0.58 to 0.59, whereas AMIF further increases this metric to 0.64 and achieves the highest VIF of 0.59. Its MI is also slightly higher than that of DM-FNET. Although FUSIONMAMBA achieves the highest competing SSIM of 0.70, AMIF reaches the same value while providing better information preservation and edge detail transfer. These results indicate that AMIF maintains structural consistency while providing a more balanced preservation of information and edge details. The advantage of AMIF becomes more evident on MRI--PET
fusion. Among the competing methods, SWINFUSION shows
the strongest overall performance, achieving an MI of 2.63,
a VIF of 0.79, and a $Q^{AB/F}$ of 0.74. In comparison,
AMIF improves these metrics to 2.97, 0.85, and 0.76,
respectively, while maintaining the same SSIM of 0.57.
FUSIONMAMBA, MMIF-INET, and DM-FNET also preserve
more complementary information than IFCNN, U2FUSION,
and PSLPT, but their MI and VIF remain clearly lower than
those of AMIF. This indicates that AMIF better preserves
complementary information from MRI and PET without
sacrificing structural consistency. A similar trend is observed on MRI--SPECT fusion. SWINFUSION again provides the strongest competing results, particularly in information preservation and structural similarity. Compared with SWINFUSION, AMIF improves MI from 2.23 to 2.68, VIF from 0.84 to 0.93, and $Q^{AB/F}$ from 0.76 to 0.78, while maintaining the same SSIM of 0.64. MMIF-INET and FUSIONMAMBA also achieve relatively competitive $Q^{AB/F}$ and SSIM values, but remain behind AMIF in information preservation and perceptual fidelity. Compared with IFCNN, U2FUSION, and PSLPT, AMIF demonstrates more pronounced advantages across most metrics. Overall, AMIF consistently achieves superior information preservation and edge detail transfer while maintaining structural similarity comparable to the best-performing methods. These results further demonstrate that integrating copyright protection and key-based authorized recovery does not compromise the fusion capability of AMIF, particularly for multimodal fusion tasks involving functional imaging modalities.\\
\textbf{Qualitative Analysis.}
As shown in Fig.~\ref{fig:AuthorizedFusionResults}, AMIF recovers watermark-free fusion results using a valid key without introducing obvious recovery artifacts. The enlarged regions further reveal differences among the compared methods in local detail preservation and cross-modal information balance. For MRI--CT fusion, some methods, such as U2FUSION and PSLPT, tend to produce smoother local regions, which weakens fine textures and structural details. In contrast, stronger competitors such as SWINFUSION and FUSIONMAMBA preserve more high-density CT information, but their results show relatively strong contrast enhancement in some regions, which can reduce the visibility of MRI soft-tissue details. MMIF-INET and DM-FNET achieve a better balance between the two modalities, although some local structures remain less distinct. AMIF preserves clear CT structural contours while retaining richer MRI soft-tissue textures, providing a more balanced representation of the complementary information. For MRI--PET and MRI--SPECT fusion, a similar difference can be observed. Some methods exhibit local smoothing or weakened anatomical textures, while the stronger competitors, including SWINFUSION, FUSIONMAMBA, MMIF-INET, and DM-FNET, preserve both anatomical and functional information more effectively. Among them, SWINFUSION shows particularly strong visual contrast and structural clarity. However, the enlarged regions indicate that AMIF retains finer anatomical textures and clearer local boundaries while maintaining the spatial distribution of the functional information. This visual advantage is consistent with the higher MI, VIF, and $Q^{AB/F}$ values reported in Table~\ref{table:Aufusionquantresults}. Overall, the qualitative comparison shows that AMIF avoids both excessive local smoothing and over-emphasis of a single source modality. Compared with the strongest competing methods, AMIF provides a more balanced preservation of structural details and complementary cross-modal information. More importantly, these properties are retained after authorized recovery, indicating that the proposed copyright protection and recovery process does not introduce noticeable degradation into the final fusion results.\\
\begin{figure}
	\centering
	\includegraphics[width=1\linewidth]{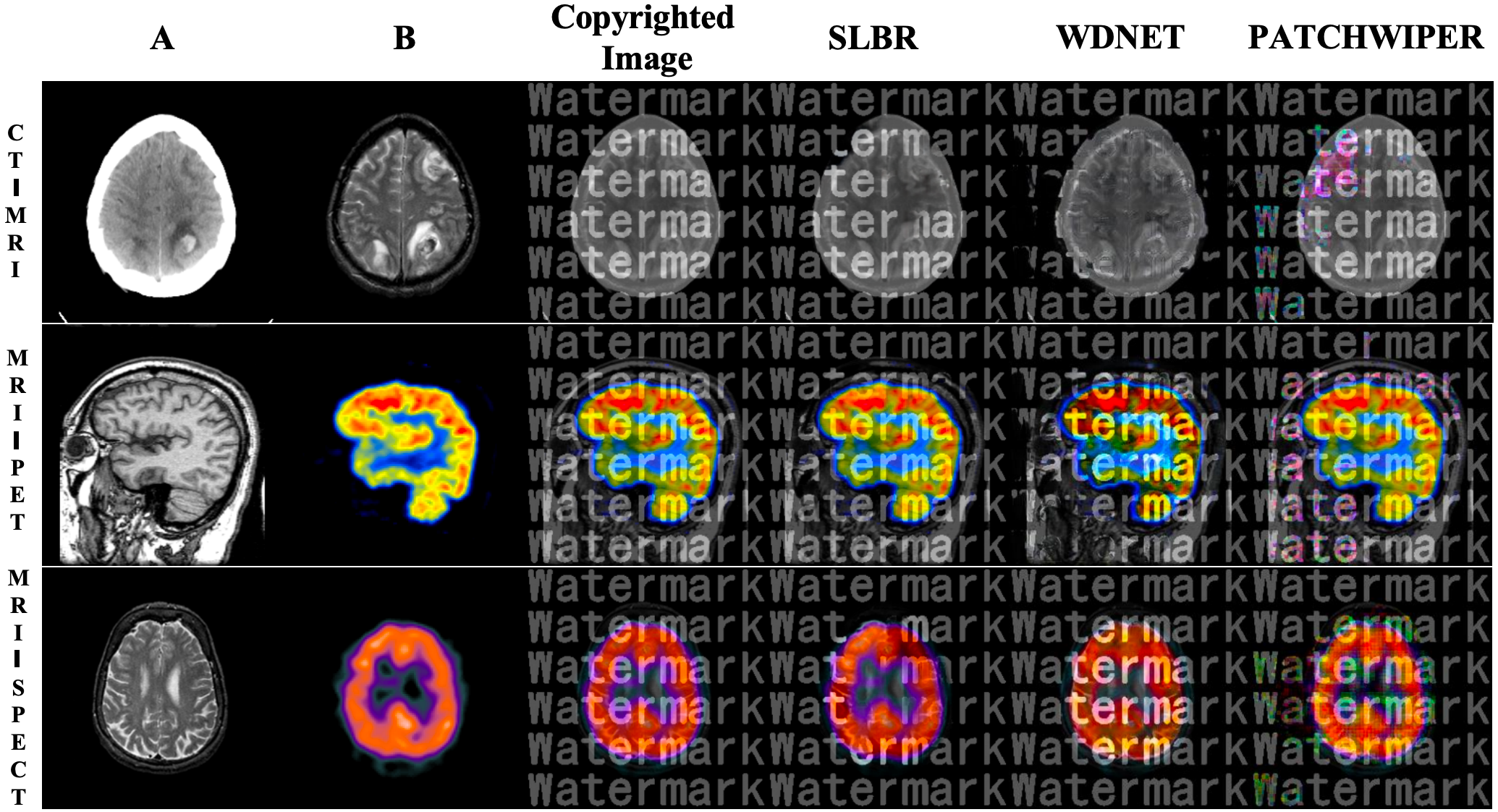}
	\caption{Visual results under watermark removal attacks. Columns A and B show the two input modalities, followed by the copyrighted fused image and the attacked outputs of SLBR, WDNET, and PATCHWIPER. Forced watermark removal in the unauthorized setting introduces distortions and artifacts, degrading the fused content.}
	\label{fig:waterAttack}
    \vspace{-10pt}
\end{figure} 
\vspace{-18pt}
\subsection{Unauthorized Watermark Removal Attacks.}
\begin{figure}
    \vspace{-10pt}
	\centering
	\includegraphics[width=0.85\linewidth]{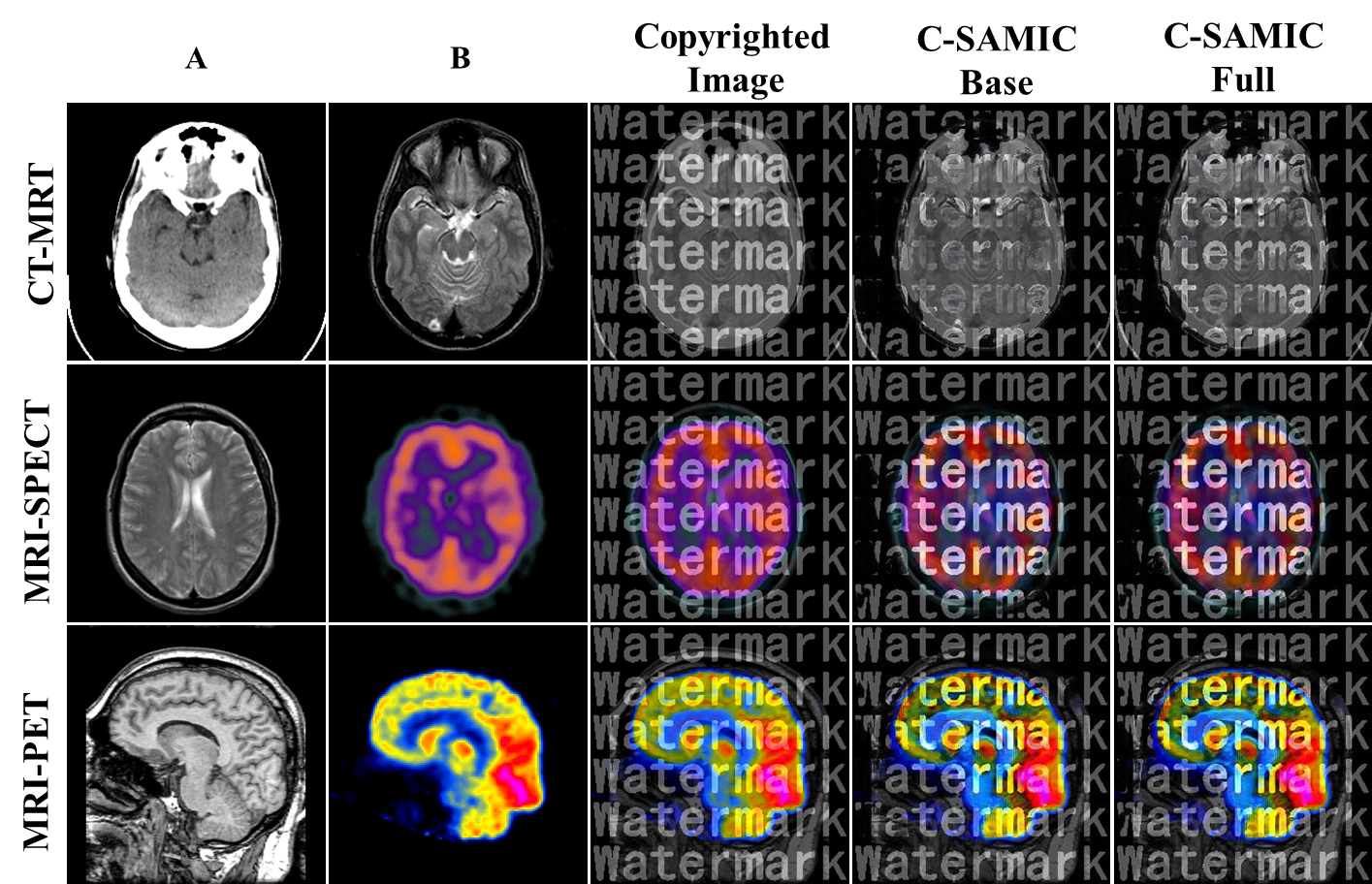}
	\caption{Ablation visualization of C-SAMIC and C-SAMIC Base under the WDNet
watermark-removal attack on the MRI--CT, MRI--PET, and MRI--SPECT test sets.
Representative examples are shown for all three modality pairs.}
	\label{fig:C-SAMICablation}
    \vspace{-13pt}
\end{figure}

Watermark removal attacks aim to suppress or remove copyright identifiers while preserving the underlying fusion content as much as possible. Therefore, the robustness of copyright protection depends not only on whether the watermark can be removed, but also on whether the attack must introduce substantial modifications to the protected fusion result. Based on this consideration, we apply multiple watermark removal methods to the copyright-protected outputs of AMIF and analyze the resulting output changes and watermark residues. This allows us to evaluate the coupling between copyright information and fusion content, as well as the resistance to unauthorized watermark removal. We adopt three representative SOTA methods for visible watermark removal, including PATCHWIPER~\cite{mo2025patchwiper}, SLBR~\cite{liang2021visible}, and WDNET~\cite{liu2021wdnet}. We apply them to the watermarked fused images produced in the unauthorized mode. \\
\textbf{Quantitative Analysis.} As shown in Table~\ref{tab:attack_protected_ssim}, different watermark removal
methods exhibit different attack strengths. Considering the full C-SAMIC
configuration, SLBR yields the highest SSIM on all three modality combinations,
with values of 0.9755, 0.9662, and 0.9832 on MRI--CT, MRI--PET, and MRI--SPECT,
respectively. This indicates that SLBR causes relatively limited changes to the
protected outputs. PatchWiper produces larger deviations, particularly on
MRI--SPECT, where SSIM decreases to 0.8742. WDNet causes the strongest
modification on MRI--CT and MRI--PET, reducing SSIM to 0.8356 and 0.8093,
respectively. These results show that the effectiveness of watermark removal
varies across both attack methods and modality combinations. \\
\textbf{Qualitative Analysis.} As shown in Fig.~\ref{fig:waterAttack}, our watermark remains clearly recognizable under these attacks, with only local corruption or smoothing. Meanwhile, the attacked results, especially the fused outputs, exhibit noticeable information loss. In regions where the watermark is weakened or partially removed, the medical content degrades significantly. Typical artifacts include damaged edges, over-smoothed tissue textures, and distorted shape or contrast in salient functional or metabolic regions. More specifically, SLBR refines the background through progressive restoration, but it usually suppresses only part of the watermark patterns on AMIF outputs. WDNET attempts to separate watermark components via decomposition, but it often causes stronger global de-texturing in this setting. This leads to reduced contrast and flattened details, while the watermark is still not cleanly removed. For AMIF, where watermark signals are deeply coupled with the content in the wavelet domain, PATCHWIPER is notably limited and may introduce additional local color abnormalities. These results indicate a strong coupling between the watermark and fused content in the feature space. Without the key, an attacker can hardly remove the watermark without damaging critical medical information, which effectively discourages unauthorized deletion.\\
\textbf{Ablation Analysis.}
For C-SAMIC, we compare the full model with a base variant that removes
$\delta$ and $\eta$ from the coupling functions. The quantitative comparison
is conducted on the MRI--CT, MRI--PET, and MRI--SPECT test sets under
PatchWiper~\cite{mo2025patchwiper}, SLBR~\cite{liang2021visible}, and
WDNet~\cite{liu2021wdnet} watermark-removal attacks. As shown in
Table~\ref{tab:attack_protected_ssim}, the full C-SAMIC consistently achieves
higher SSIM than the Base variant under all three watermark removal attacks
and across all three modality combinations. As shown in Fig.~\ref{fig:C-SAMICablation}, C-SAMIC keeps the watermark clear
and continuous after the WDNet attack, while the protected content shows only
mild degradation. In contrast, the base variant exhibits weakened watermark
patterns and stronger content damage, such as over-smoothed details and
degraded anatomical boundaries. These results suggest that functions
$\delta$ and $\eta$ strengthen feature interaction in both channel and spatial
dimensions, enabling tighter coupling between the watermark and structural
content. This improves watermark embedding strength and robustness against
unauthorized watermark removal.
\begin{table}[t]
\centering
\caption{Robustness comparison of the C-SAMIC ablation configurations under different
watermark removal attacks on three multimodal test sets. SSIM measures the similarity
between each attacked output and its corresponding copyright-protected fusion image.
Higher values indicate that the attacked output remains closer to the protected fusion
image, corresponding to stronger resistance to attack-induced modification
(\textbf{bold} indicates the best result).}
\label{tab:attack_protected_ssim}

\begingroup
\scriptsize
\setlength{\tabcolsep}{2.5pt}
\renewcommand{\arraystretch}{0.92}

\begin{tabular}{@{}llccc@{}}
\toprule
\bfseries Attack & \bfseries C-SAMIC
& \bfseries MRI--CT
& \bfseries MRI--PET
& \bfseries MRI--SPECT \\
& &
\bfseries SSIM$\uparrow$
& \bfseries SSIM$\uparrow$
& \bfseries SSIM$\uparrow$ \\
\midrule

PatchWiper~\cite{mo2025patchwiper}
& Base & 0.8870 & 0.9073 & 0.8681 \\
& Full & \textbf{0.9002} & \textbf{0.9137} & \textbf{0.8742} \\

\midrule

SLBR~\cite{liang2021visible}
& Base & 0.9505 & 0.9368 & 0.9679 \\
& Full & \textbf{0.9755} & \textbf{0.9662} & \textbf{0.9832} \\

\midrule

WDNet~\cite{liu2021wdnet}
& Base & 0.7351 & 0.7273 & 0.8323 \\
& Full & \textbf{0.8356} & \textbf{0.8093} & \textbf{0.9315} \\

\bottomrule
\end{tabular}

\endgroup
\end{table}
\vspace{-10pt}
\subsection{Comparison with Visible Watermark Protection Methods under Removal Attacks.}
To evaluate the comparative robustness of AMIF against unauthorized watermark removal, we compare it with conventional alpha blending~\cite{tsai2010secure} and the attack-aware RIE watermark protection method~\cite{RIE}.
For Alpha blending and RIE, we first generate watermark-free reference fusion results by disabling the copyright protection module while keeping the same AMIF framework and multimodal inputs. These reference fusion images are then used as the clean inputs for subsequent visible watermark protection. For a fair comparison, we adjust the blending strength of Alpha such that the Mean Squared Error (MSE) introduced by watermark embedding is approximately matched to that introduced by AMIF. For RIE, we follow its original attack-aware protocol and generate protected outputs separately for each removal model. The protected outputs generated by all methods are subsequently attacked using three representative visible watermark removal models, including PatchWiper~\cite{mo2025patchwiper}, WDNet~\cite{liu2021wdnet}, and SLBR~\cite{liang2021visible}.

Table~\ref{tab:watermark_removal_ncc} reports the normalized cross-correlation (NCC) of the residual watermark after different watermark-removal attacks, where a higher NCC indicates better preservation of the original watermark structure and thus stronger resistance to watermark removal. AMIF achieves the best NCC in eight of nine settings. It consistently outperforms Alpha and RIE under all three attacks on MRI--CT and MRI--PET, and remains superior under PatchWiper and SLBR on MRI--SPECT. The only exception occurs for MRI--SPECT under WDNet, where Alpha achieves an NCC of 0.79 compared with 0.76 for AMIF. Averaged over all nine settings, AMIF obtains an NCC of 0.82, outperforming Alpha (0.74) and RIE (0.57). Notably, RIE requires attack-specific optimization for different removal models, whereas AMIF uses a unified protected output without prior knowledge of the attack model. Nevertheless, the original watermark structure remains well preserved after attack.
\begin{table}[t]
\centering
\caption{Quantitative comparison of NCC for different visible
watermark protection methods under three watermark-removal
attacks. NCC is computed only within the watermark region.
Higher values indicate better preservation of the original
watermark structure after attack. \textbf{Bold} indicates
the best result.}
\label{tab:watermark_removal_ncc}

\begingroup
\scriptsize
\setlength{\tabcolsep}{2pt}
\renewcommand{\arraystretch}{0.95}

\begin{tabular}{@{}llccc@{}}
\toprule
\textbf{Dataset}
& \textbf{Attack}
& \textbf{Alpha}~\cite{tsai2010secure}
& \textbf{RIE}~\cite{RIE}
& \textbf{AMIF (OURS)} \\
\midrule

\multirow{3}{*}{MRI--CT}
& PatchWiper & 0.59 & 0.60 & \textbf{0.85} \\
& WDNet      & 0.77 & 0.73 & \textbf{0.84} \\
& SLBR       & 0.87 & 0.75 & \textbf{0.96} \\
\midrule

\multirow{3}{*}{MRI--PET}
& PatchWiper & 0.71 & 0.72 & \textbf{0.74} \\
& WDNet      & 0.68 & 0.55 & \textbf{0.77} \\
& SLBR       & 0.85 & 0.55 & \textbf{0.91} \\
\midrule

\multirow{3}{*}{MRI--SPECT}
& PatchWiper & 0.51 & 0.53 & \textbf{0.63} \\
& WDNet      & \textbf{0.79} & 0.37 & 0.76 \\
& SLBR       & 0.87 & 0.36 & \textbf{0.95} \\
\bottomrule

\end{tabular}
\endgroup
\end{table}
\vspace{-10pt}
\subsection{Authorization-Key Validity Analysis}
\begin{figure}
    \centering
    \includegraphics[width=1\linewidth]{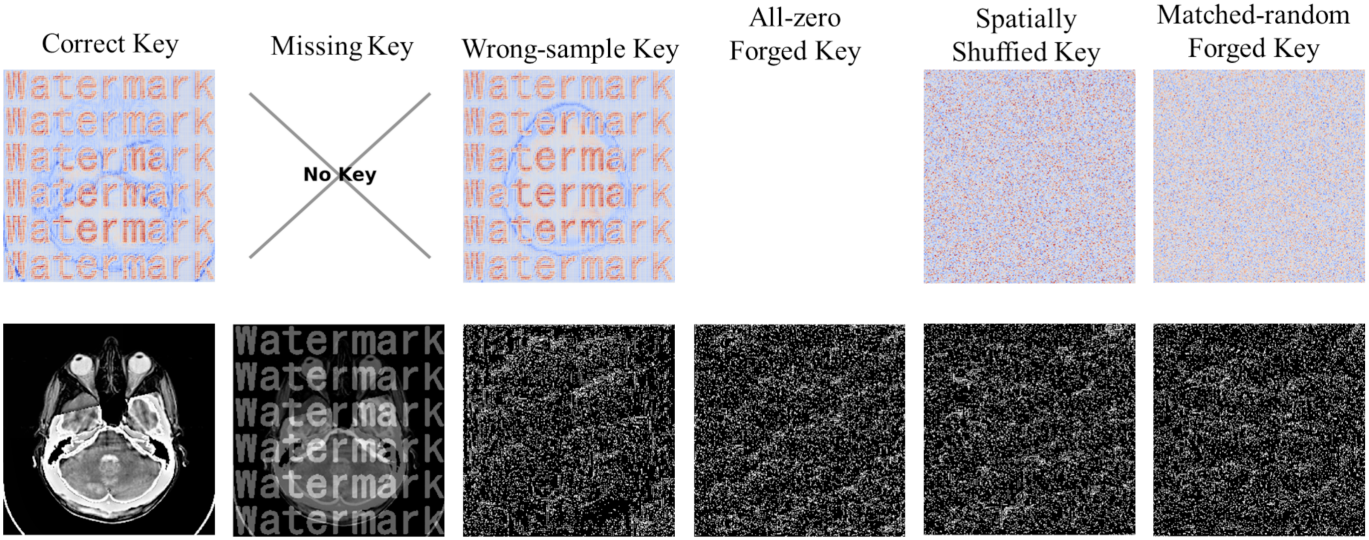}
    \caption{Qualitative comparison under different authorization-key conditions. 
    For visualization convenience, the keys in the top row are displayed using a red--blue colormap, while the bottom row shows the corresponding recovery results.}
    \label{fig:differentkeysvisualeffects}
    \vspace{-18pt}
\end{figure}

To comprehensively evaluate the role of the authorization key under different usage scenarios, we examine two complementary aspects: key correctness and key integrity. First, for unauthorized or invalid-access scenarios, we evaluate the recovery process using the correct key, a missing key, a key from another sample, and several forged keys. This experiment examines whether high-fidelity recovery can only be achieved with the valid key matched to the current sample. We further evaluate the valid key under controlled perturbations. Gaussian noise and quantization are used to simulate numerical errors and precision degradation that may occur during key storage, transmission, or processing, while channel erasure is used to assess authorized recovery when part of the key information is missing. The detailed results of key-integrity evaluation are provided in the S.II-C. To independently evaluate recovery fidelity, watermark-free fusion outputs are generated by disabling the copyright protection module while keeping the same AMIF framework and multimodal inputs. The outputs obtained under different key conditions are then compared with their corresponding watermark-free fusion references using MI, VIF, and SSIM. As shown in Table~\ref{tab:recovery-information-correctness}, only the correct authorization key enables high-fidelity recovery of the watermark-free fusion result. Across all three test sets, the recovered images achieve SSIM values close to 1.0, indicating nearly lossless authorized recovery. When the key is missing, recovery quality drops substantially. Using a key from another sample leads to further degradation. For all-zero, spatially shuffled, and statistically matched random forged keys, SSIM values approach zero. These results demonstrate that high-fidelity recovery strongly depends on the correct sample-specific key. Fig.~\ref{fig:differentkeysvisualeffects} further visualizes the recovery results under different authorization-key conditions. With the correct key, AMIF successfully recovers clean watermark-free fusion images with clear anatomical structures and no visible copyright residues. When the key is missing, authorized recovery cannot be activated, and the model retains the copyright-protected output with the visible watermark. In contrast, the wrong-sample key fails to recover the corresponding fusion content and introduces severe structural distortions. Similarly, the all-zero, spatially shuffled, and matched-random forged keys produce highly degraded outputs dominated by irregular noise and artifacts, with little meaningful anatomical structure preserved. These visual observations are consistent with the quantitative results in Table~\ref{tab:recovery-information-correctness}, further demonstrating that valid recovery is specifically tied to the correct sample-matched authorization key.
\begin{table}[t]
\centering
\caption{Quantitative evaluation of authorization-key correctness on three
multimodal test sets (\textbf{bold} indicates the best fidelity).}
\label{tab:recovery-information-correctness}

\begingroup
\tiny
\setlength{\tabcolsep}{0.8pt}
\renewcommand{\arraystretch}{0.95}

\resizebox{\columnwidth}{!}{%
\begin{tabular}{lccc ccc ccc}
\toprule
\bfseries Authorization Key
& \multicolumn{3}{c}{\bfseries MRI--CT}
& \multicolumn{3}{c}{\bfseries MRI--PET}
& \multicolumn{3}{c}{\bfseries MRI--SPECT} \\
\cmidrule(lr){2-4}\cmidrule(lr){5-7}\cmidrule(lr){8-10}

& \bfseries MI $\uparrow$ & \bfseries VIF $\uparrow$ & \bfseries SSIM $\uparrow$
& \bfseries MI $\uparrow$ & \bfseries VIF $\uparrow$ & \bfseries SSIM $\uparrow$
& \bfseries MI $\uparrow$ & \bfseries VIF $\uparrow$ & \bfseries SSIM $\uparrow$ \\

\midrule
Correct Key
& \textbf{2.9238} & \textbf{1.0000} & \textbf{0.999988}
& \textbf{3.5081} & \textbf{1.0000} & \textbf{0.999998}
& \textbf{2.8278} & \textbf{1.0001} & \textbf{0.999998} \\

Missing Key
& 1.1186 & 0.0813 & 0.271102
& 1.3801 & 0.0709 & 0.277403
& 1.2727 & 0.0669 & 0.258948 \\

Wrong-sample Key
& 0.0046 & 0.0153 & 0.110024
& 0.0792 & 0.0126 & 0.011268
& 0.0116 & 0.0218 & 0.100447 \\

All-zero Forged Key
& 0.0020 & 0.0129 & 0.002313
& 0.0021 & 0.0118 & 0.003303
& 0.0020 & 0.0155 & 0.002865 \\

Spatially Shuffled Key
& 0.0021 & 0.0128 & 0.002567
& 0.0021 & 0.0115 & 0.003154
& 0.0020 & 0.0157 & 0.002999 \\

Matched-random Forged Key
& 0.0021 & 0.0126 & 0.002461
& 0.0021 & 0.0116 & 0.003057
& 0.0020 & 0.0158 & 0.003006 \\

\bottomrule
\end{tabular}%
}
\endgroup
\vspace{-10pt}
\end{table}
\vspace{-10pt}
\subsection{CCWM Ablation Study}
\begin{table}[t]
    \centering
    \caption{Quantitative ablation comparison of AMIF with and without CCWM on three multimodal test sets (\textbf{bold} indicates the best, W/ and W/O denote with and without, respectively).}
    \label{table:ccwmxiaorong}

    \setlength{\tabcolsep}{3pt}
    \renewcommand{\arraystretch}{1.15}
    \footnotesize

    \begin{tabular}{clcccc}
        \toprule
        \textbf{Dataset}
        & \textbf{Variant}
        & \textbf{MI} $\uparrow$
        & \textbf{VIF} $\uparrow$
        & $\mathbf{Q^{AB/F}}$ $\uparrow$
        & \textbf{SSIM} $\uparrow$ \\
        \midrule

        \multirow[c]{2}{*}{MRI-CT}
        & W/O CCWM & 1.99 & 0.57 & \textbf{0.64} & \textbf{0.70} \\
        & W/ CCWM  & \textbf{2.04} & \textbf{0.59} & \textbf{0.64} & \textbf{0.70} \\
        \midrule

        \multirow[c]{2}{*}{MRI-PET}
        & W/O CCWM & 2.91 & 0.84 & \textbf{0.76} & \textbf{0.57} \\
        & W/ CCWM  & \textbf{2.97} & \textbf{0.85} & \textbf{0.76} & \textbf{0.57} \\
        \midrule

        \multirow[c]{2}{*}{MRI-SPECT}
        & W/O CCWM & 2.62 & 0.91 & \textbf{0.78} & 0.63 \\
        & W/ CCWM  & \textbf{2.68} & \textbf{0.93} & \textbf{0.78} & \textbf{0.64} \\

        \bottomrule
    \end{tabular}
    \vspace{-10pt}
\end{table}
The ablation experiments use the same training protocol, hyper-parameters, and test set as the authorized fusion experiments reported in Sec.~4.3. To ablate CCWM, we replace the full module with a simplified design that retains only the learnable watermark parameters, while leaving all other modules unchanged. Table~\ref{table:ccwmxiaorong} shows consistent drops across all three datasets after removing CCWM. The decrease is more evident on MI and VIF, which reflect information preservation and perceptual fidelity. This indicates that CCWM conditions watermark representations on the input content. The resulting watermark features implicitly encode multimodal cues. This provides more informative content signals during watermark embedding and key-driven recovery. It improves watermark-free reconstruction quality and cross-modal information retention.
\vspace{-18pt}
\section{Conclusion}
We present AMIF, the first authorizable medical image fusion model with built-in authentication. AMIF unifies multimodal medical image fusion, copyright protection, and key-based authorization within a single framework, making copyright protection an intrinsic capability of the fusion model. In the unauthorized mode, AMIF generates fused images with explicit visible copyright identifiers, while a valid authorization key enables high-quality watermark-free recovery. To support this framework, CCWM generates content-conditioned watermark representations from internal learnable parameters and incorporates implicit multimodal content cues, while C-SAMIC establishes tighter coupling between watermark information and protected fusion features through channel--spatial modulated invertible interaction. Experiments on MRI--CT, MRI--PET, and MRI--SPECT datasets demonstrate that AMIF achieves competitive fusion quality after authorized recovery, supports reliable key-dependent authorization, and exhibits strong robustness against representative visible watermark removal attacks and model-parameter perturbations. Future work could further explore tighter coupling between built-in copyright protection mechanisms, fusion model architectures, and image content, allowing copyright protection to be jointly optimized with the image fusion process as an intrinsic model capability. The generalizability and scalability of the authorizable fusion paradigm could also be investigated across different fusion architectures and broader multimodal fusion tasks, together with more flexible authorization mechanisms such as multi-user authorization and hierarchical access control.

\bibliographystyle{splncs04}
\bibliography{main}

\end{document}